\documentclass[journal]{IEEEtran}
\usepackage[T1]{fontenc}
\usepackage{graphicx}
\usepackage{graphicx}
\usepackage{times}
\usepackage{helvet}
\usepackage{courier}
\usepackage{amsmath}
\usepackage{algorithm}
\usepackage{algorithmic}
\usepackage{csquotes} 
\usepackage{color}
\usepackage{paralist}
\usepackage{amssymb}
\usepackage{indentfirst}
\usepackage{subfigure}
\usepackage{float}
\usepackage{multirow}
\usepackage{cite}
\usepackage{mathrsfs}

\usepackage{mathrsfs}
\usepackage[colorlinks, linkcolor=red,  anchorcolor=blue, citecolor=blue]{hyperref}
\usepackage{textcomp,booktabs}
\usepackage{amssymb}
\usepackage{pifont}

\usepackage{makecell}

\begin{document}

\title{MFGNet: Dynamic Modality-Aware Filter Generation for RGB-T Tracking}

\author{Xiao Wang, Xiujun Shu, Shiliang Zhang, Bo Jiang, Yaowei Wang, \emph{Member, IEEE} \\ Yonghong Tian, \emph{Fellow, IEEE}, Feng Wu, \emph{Fellow, IEEE} 
\thanks{Xiao Wang, Xiujun Shu, Shiliang Zhang, Yaowei Wang, and Feng Wu are with Peng Cheng Laboratory, Shenzhen, China. Shiliang Zhang and Yonghong Tian are also with Peking University, Beijing, China. Bo Jiang is with the School of Computer Science and Technology, Anhui University, Hefei, China. Feng Wu is also with University of Science and Technology of China, Hefei, China.  
Corresponding author: Yaowei Wang.  Email: wangxiaocvpr@foxmail.com, \{shuxj, wangyw, tianyh\}@pcl.ac.cn, slzhang.jdl@pku.edu.cn, jiangbo@ahu.edu.cn, fengwu@ustc.edu.cn.}}

\markboth{IEEE Transactions on Multimedia}  
{Shell \MakeLowercase{\textit{et al.}}: Bare Demo of IEEEtran.cls for IEEE Journals}

\maketitle

\begin{abstract}
Many RGB-T trackers attempt to attain robust feature representation by utilizing an adaptive weighting scheme (or attention mechanism). Different from these works, we propose a new dynamic modality-aware filter generation module (named MFGNet) to boost the message communication between visible and thermal data by adaptively adjusting the convolutional kernels for various input images in practical tracking. Given the image pairs as input, we first encode their features with the backbone network. Then, we concatenate these feature maps and generate dynamic modality-aware filters with two independent networks. The visible and thermal filters will be used to conduct a dynamic convolutional operation on their corresponding input feature maps respectively. Inspired by residual connection, both the generated visible and thermal feature maps will be summarized with input feature maps. The augmented feature maps will be fed into the RoI align module to generate instance-level features for subsequent classification. To address issues caused by heavy occlusion, fast motion and out-of-view, we propose to conduct a joint local and global search by exploiting a new direction-aware target driven attention mechanism. The spatial and temporal recurrent neural network is used to capture the direction-aware context for accurate global attention prediction. Extensive experiments on three large-scale RGB-T tracking benchmark datasets validated the effectiveness of our proposed algorithm.  The source code of this paper is available at \textcolor{magenta}{\url{https://github.com/wangxiao5791509/MFG_RGBT_Tracking_PyTorch}}. 
\end{abstract}


\IEEEpeerreviewmaketitle

\section{Introduction}
\IEEEPARstart{O}{bject}  tracking is a popular research topic in computer vision that aims to locate a determined object (initialized in the first frame) in each video frame. It has been widely used in many applications, such as intelligent surveillance, automatic driving, and unmanned aerial vehicles. Although it has already achieved great success in recent years with robust target representation brought by deep neural network \cite{wang2018DAT, wang2018sint++, Nam2015Learning, wang2018learningAttention, bertinetto2016fully, danelljan2017eco, yuan2015visual, liu2017visualTrack, liu2017robustTrack, wang2018learningAtt, yao2016real, hu2019robust, zhang2015object, zhang2021learn, zhu2022TOT}, these trackers still suffer from challenging factors, e.g., illumination, scale variation, and fast motion.

\begin{figure*}[!htb]
\center
\includegraphics[width=7in]{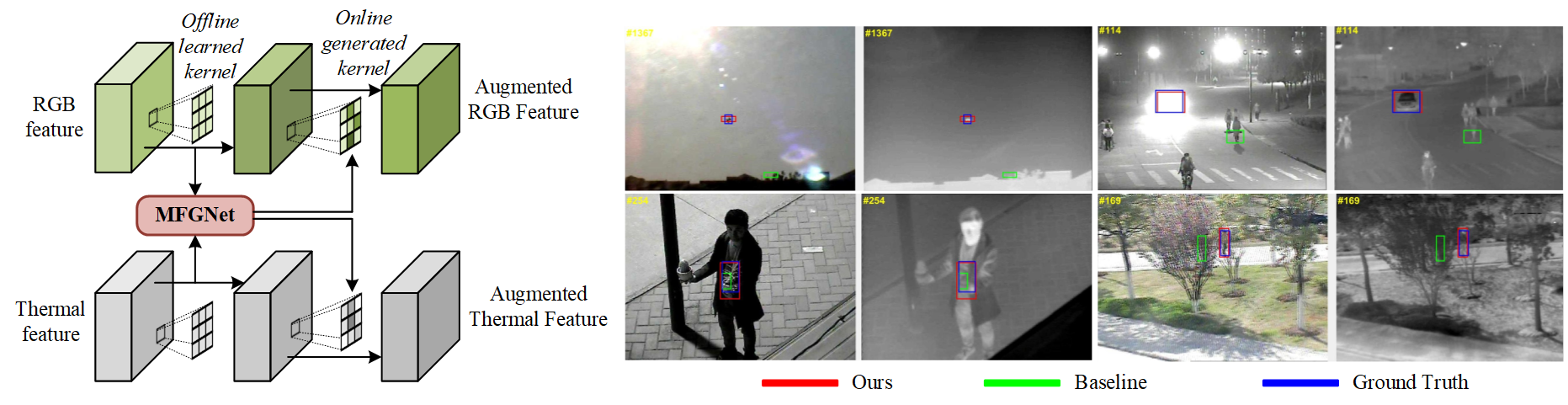}
\caption{\textcolor{black}{a). Illustration of our dynamic modality-aware filter generation module; b). Comparison between tracking results of regular CNN and our MFGNet.}}  
\label{frontImage}
\end{figure*}

To handle these issues, many researchers resort to multi-modal data to improve the performance of visual tracking \cite{wang2018DAT, li2019rgb, wang2021visevent, lukezic2019cdtb, yang2021rgbt}. Among them, RGB-T tracking has attracted more and more attention in recent years \cite{wang2019qualityDRL, li2017weightedRGBT210, zhu2018fanet, Li2018Learning, LiTwo, li2019rgb, Wang_2020_CMPP} which has great potential to work on all-time and all-weather. Existing multi-modal based algorithms attempt to model the relations between dual modalities using element-wise addition \cite{chen2019multi}, concatenation \cite{gao2019DAFNet} and convolution operations \cite{chen2018progressively}, or considering the quality of different data \cite{LiTwo, li2019rgb, CrossLi2018, li2018fusing, zhu2018fanet}, suppressing the noise in the bounding box \cite{CrossLi2018, LiTwo, li2019rgb} or selecting useful features \cite{li2018fusing}. Some RGB-T trackers utilize attention mechanisms to learn different modality weights and achieve better tracking results \cite{yang2019learning}. However, their performance in challenging scenarios is still unsatisfactory due to the influence of thermal crossover, clutter background, etc. How to fuse the two modalities adaptively for robust RGB-T tracking is still an open problem.

According to our observation, we find that the standard CNN used in previous RGB-T trackers adopts \emph{static} convolutional operation with the kernels learned offline. In another word, their convolutional kernels used in the tracking phase are content-agnostic and fixed after training. As noted in many works \cite{jia2016dynamicCFilter, wu2018dynamic, diba2019dynamonet, kang2017incorporating, shen2018DFN, gong2018CIN, liu2020learning, pang2020hierarchical, xu2020unified}, this operation can't process the test data with \emph{dynamic filters} (i.e., the specific convolutional kernels predicted for each input using a filter generation module in an online manner) which may lead to sub-optimal results. In contrast, the dynamic convolutions are content-adaptive and adapt to different inputs even after training. This may indicate that dynamic convolutions can handle challenging tracking task more suitable than standard convolutions.  These observations all inspire us to design new modules for RGB-T tracking.

\begin{figure*}[!htb]
\center
\includegraphics[width=5in]{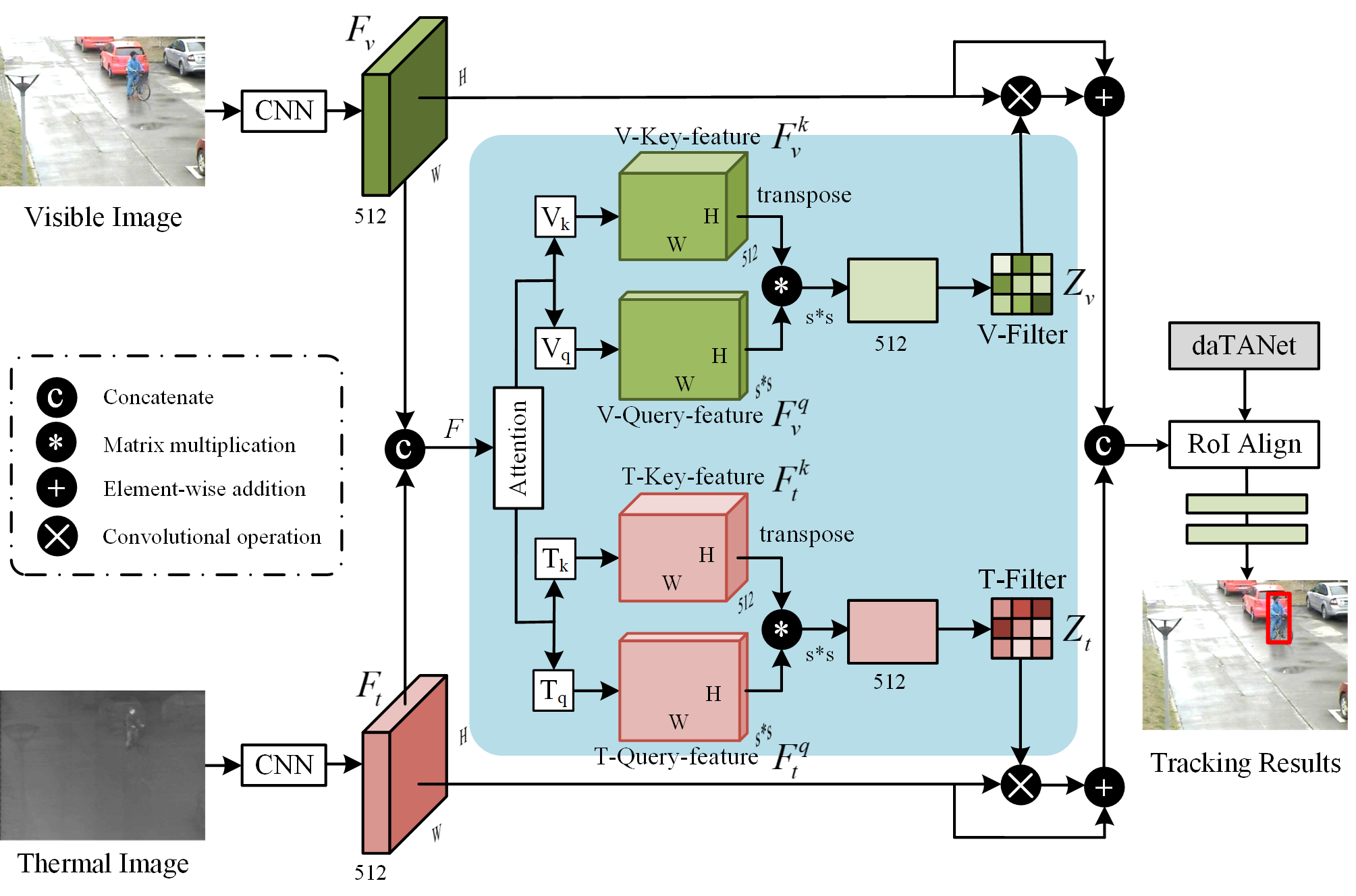}
\caption{Our modality-aware dynamic filter generation for RGB-T tracking. For the details about the daTANet, please refer to Fig. \ref{RGBT_AttentionModule}. }
\label{pipeline}
\end{figure*}

With this in mind, we propose a novel dynamic Modality-aware Filter Generation Network (named MFGNet) to boost the communication between dual-modalities for RGB-T tracking. Our network can adaptively adjust the convolutional filters for various input images in practical tracking, as shown in Fig. \ref{frontImage}. These online predicted kernels can make more efficient convolution operations for current RGB and thermal features and promote the network to obtain more flexible and targeted features for RGB-T tracking. In more detail, the features of each modality are first encoded with a shared backbone network. Then, these feature maps are concatenated together and fed into two independent dynamic modality-aware filter generation networks to predict modality-specific kernels. We conduct dynamic convolutional operation with the obtained filters on feature maps of each modality independently to attain modality-specific features. It is worth mentioning that our model can also be utilized together with attention schemes, like spatial and channel attention, for robust feature learning. In this paper, we introduce such attention mechanisms into our pipeline as shown in Fig. \ref{pipeline} to further augment our features.

With the aforementioned features, we can directly train an online classifier and conduct tracking in the tracking-by-detection framework, as many RGB-T trackers do \cite{zhu2019dense, yang2019learning}. The trackers developed based on this framework adopt local search and this makes their trackers sensitive to challenging factors, as noted in \cite{wang2019GANTrack, yang2019learning, wang2021ganTANet, wang2021TNL2K, wang2021deepmta}. Therefore, they design the target-driven global attention scheme for the joint local and global search to handle this issue. Based on their model, we propose to analyze the spatial and temporal context in a \emph{direction-aware} manner to have a full understanding of global image semantics and motions for better attention prediction. An overview of our direction-aware target driven attention network (termed daTANet) can be found in Fig. \ref{RGBT_AttentionModule}.

We integrate our proposed modules into real-time mdnet (RT-MDNet) \cite{jung2018realMDNet} and also conduct extensive experiments on three large-scale RGB-T tracking benchmark datasets to validate the effectiveness of our proposed algorithm. Specifically, we achieve 0.749/0.494 and 0.780/0.535 (PR/SR) on the RGBT-210 and RGBT-234 datasets, outperforming the baseline method by +3.4/2.7, +2.5/2.0 respectively. We also outperform the baseline approach by a large margin +8.6/5.4 on the GTOT-50 dataset (Ours: 0.889/0.707, baseline: 0.803/0.653). These experiments all validate the advantages and effectiveness of our proposed modules for RGB-T tracking.

The contributions of this paper can be summarized as: 
 
1). We propose a novel dynamic modality-aware filter generation network for RGB-T tracking. To the best of our knowledge, this paper is the first work to connect dual modalities via dynamic filter prediction. We also validate that our model can work well with spatial and channel attention for more robust RGB-T tracking. 

2). We introduce the direction-aware target driven attention network for global search which can further improve the final tracking performance.

\section{Related Work} 
In this section, we will give a brief review of the RGB-T tracking, dynamic filter generation, visual attention, \textcolor{black}{and spatio-temporal feature learning. More related works on the visual trackers and dynamic neural networks can be found in the following survey papers \cite{zhang2020multi, zhang2020RGBTSurvey, marvasti2021deepsurvey, han2021DNNsurvey} and paper list \footnote{\url{https://github.com/wangxiao5791509/RGB-Thermal-Tracking-Paper-List}}. }

\textbf{RGB-T Tracking.} 
\textcolor{black}{Many existing trackers are developed based on RGB cameras, including binary classification based, siamese matching based, and correlation filter based tracking. Note that, Chen et al. \cite{chen2016robust, chen2015real} propose a discriminative patch-based appearance model for robust visual tracking based on compressive sensing. Zhong et al. \cite{zhong2018hierarchical} propose a coarse-to-fine tracking framework, more in detail, the data-driven search scheme is used for coarse-level target localization and the fine-level verification is then conducted for more accurate tracking. Zhou et al. \cite{zhou2018deepalignNet} propose a deep alignment network based tracker to address the misalignment in the pedestrian detectors by considering occlusion and motion reasoning.}   
With the popularity of thermal infrared sensors, RGB-T tracking attracted more and more attention in the field of computer vision. In recent years, there are three large-scale benchmark datasets released for this task, i.e., the GTOT-50 \cite{Li2016Learning}, RGBT-210 \cite{li2019rgb} and RGBT-234 \cite{li2019rgb}. From the perspective of algorithms, Li et al. \cite{Li2018Learning} proposes a local-global multi-graph descriptor that can suppress background effects to improve the performance of RGB-T tracking. 
Wu et al. \cite{Wu2011Multiple} directly concatenate the RGB-T image patches and use a sparse representation of each sample in the target template space for tracking. A two-stream ConvNet and a Fusion Net are proposed in \cite{li2018fusing} for RGB-T tracking. 
Li et al. \cite{LiTwo} proposes a two-stage modality-graphs regularized manifold ranking algorithm for learning a more robust representation for RGB-T tracking.  
A novel cross-modal manifold ranking algorithm that considers the inconsistency between different modalities is introduced in \cite{CrossLi2018} to reduce the noise effect. 
Zhu et al. \cite{zhu2019dense} propose a dense feature aggregation module for multi-modal and multi-layer feature fusion, in addition, a pruning net is employed to filter the useless features. 
Li et al. \cite{long2019MANet} adopts the multi-layer framework to model the modality-specific and modality-shared features. 
Zhang et al. \cite{zhang2019mfDiMP} propose an end-to-end framework for RGB-T tracking and analyses the performance of different fusing strategies, including early fusion, middle fusion, and late fusion. 
Although these RGB-T tracking algorithms have achieved good performance, they use static kernels to get their features, which may limit their final tracking performance.

\textbf{Dynamic Filter Generation.}  
Different from standard convolutional networks that learn their filters in an offline manner, the dynamic filter networks proposed by Xu et al. \cite{jia2016dynamicCFilter} can learn specific parameters for different input data. This model is powerful, flexible, and also used in many practical applications. Wu et al. \cite{wu2018dynamic} further propose the LS-DFN and learn the position-specific kernels from not only the identical position but also multiple sampled neighbor regions. He et al. \cite{he2019dynamicMSF} introduce the idea of multi-scale leaning into the dynamic filter network and improve the semantic segmentation significantly. Ali et al. \cite{diba2019dynamonet}  develop the DynamoNet which can learn the motion cues in videos in self-supervised learning with dynamic motion filters. Kang et al. \cite{kang2017incorporating} propose the ACNN to incorporate the available side information for their task based on a dynamic filter network. Shen et al. \cite{shen2018DFN} extract the intermediate feature maps from an autoencoder and learn a set of coefficients to attain their sample-specific features. In addition, some NLP related tasks also utilize the dynamic filter to model the relations between sentences \cite{shen2017learningDFN, gong2018CIN}. More works on dynamic neural networks can be found in the survey \cite{han2021DNNsurvey}. Different from these works, we are the first to utilize the dynamic filter network for the multi-modal task, i.e., the RGB-T tracking. We adopt DFN for robust feature learning to mine the modality-aware context information for robust RGB-T tracking.

\textbf{Attention Mechanism.}  
In the deep learning era, the attention model is widely used in many tasks which firstly originated from cognitive neuroscience. For the tracking task, Pu et al. \cite{pu2018deep} propose an attentive method for visual tracking which can make the tracker attend to target regions. Wang et al. \cite{wang2019GANTrack, wang2018DAT, wang2021TANettnnls} propose target-aware attention and conduct joint local and global search for visual tracking. Choi et al. propose an attentional mechanism that chooses a subset of the associated correlation filters for visual tracking in AFCN \cite{choi2017AFCN}. Wang et al. \cite{wang2018attentionTrack} introduce multiple attention models to attain better features for tracking. Yang et al. \cite{yang2020siamatt} introduce an attention network in the classification branch of RPN, then discriminate the positive and negative samples by weighting-fusion of the score of the classification branch and attention branch. Shen et al. \cite{shen2019VTAttention} introduce the attention mechanism into the Siamese network for object tracking and propose a new way to compute attention weights for objects. The authors of \cite{rahman2020efficient} present stacked channel-spatial attention within a Siamese framework to learn effective feature representation and discrimination ability for high-performance tracking. Cui et al.  \cite{cui2016recurrently} utilize multi-directional RNN to predict the saliency regions for visual tracking. In this paper, we introduce the spatial and channel attention for robust feature learning and propose the direction-aware target driven attention for global search respectively.

\textcolor{black}{\textbf{Spatio-temporal Feature Learning.}  As is known to all, the spatiotemporal features are imperative for video related tasks, such as visual tracking, video saliency detection \cite{chen2017video, li2019accurate, li2020plug, chen2019improved}, action recognition, etc. Specifically, Chen et al. \cite{chen2017video} explore the spatial-temporal saliency fusion and low-rank coherency guided saliency diffusion for video saliency detection. Li et al. \cite{li2020plug} build a plug-and-play module that can introduce newly sensed and coded temporal information for video saliency detection, by weakly retraining a pre-trained image saliency deep model. The authors of work \cite{chen2019improved} attempt to firstly identify the beyondscope frames with trustworthy long-term saliency clues to address the issue of long period less-trustworthy spatial-temporal saliency clues. Li et al. \cite{li2019accurate} exploit the spatial-temporal coherency of the salient foregrounds and the objectness prior for long-term information enhanced video saliency detection. In addition, the 
} 
LSTM \cite{hochreiter1997lstm} is developed for sequential data processing which has been widely used in many applications of computer vision and natural language processing. It contains four gates, including the input gate, forget gate, memory gate, and output gate. The original LSTM adopts the fully connected layer to conduct information transformation which limits its application on image domains. ConvLSTM \cite{xingjian2015convlstm} is proposed to replace the fully connected layer with convolutional layer and many researches validated its effectiveness \cite{ye2020dualconvlstm}. LSTM is also widely used in visual tracking problem and many RNN based trackers are proposed like \cite{ondruska2016rnntrack, yang2018DMN, ondruska2016track, bhat2020know, gan2015firstrnn, ning2017spatiallyrnn, yang2017recurrent}. 
Specifically, Goutam Bhat et al. \cite{bhat2020know} propose to use ConvGRU to model the scene information for object tracking based on DiMP \cite{bhat2019DiMP}. 
Yang et al. \cite{yang2018DMN} adopt the dynamic memory network to update the target model to address the issue of appearance variation of the target object. Gan et al. \cite{gan2015firstrnn} use an RNN to regress the target locations directly. The authors of \cite{ning2017spatiallyrnn} use the object detector to obtain initial object proposals, then, use an LSTM to attain their target box. Different from these works, we adopt the LSTM to capture the spatial and temporal information for more accurate target-aware attention prediction.

\section{The Proposed Approach}

In this section, we will first define the problem formulation and introduce the motivation of this paper. Then, we give an overview of our proposed RGB-T tracker. After that, we will dive into the details of our proposed dynamic modality-aware filter generation network, and direction-aware target driven attention module. Finally, we will integrate the proposed modules with RT-MDNet and discuss the details in \textcolor{black}{the training and testing phase respectively.}

\subsection{Problem Formulation} 
Given the $i$-th input RGB and thermal image pairs $I_v^i, I_t^i$, the RGB-T tracking task is formulated as a binary classification problem in this paper, i.e., discriminate the given $N$ proposal $(x_{1, i}, x_{2, i}, ... , x_{N, i} )$ is foreground or background \cite{Jung_2018_ECCV, Park_2018_ECCV, Nam2015Learning}. Usually, we select the proposal with maximum classification score: 
\begin{equation}
\label{argMaxFunction}
\hat{x}_{j, i} = \mathop{\arg\max}_{x_{j, i}} ~  \mathop{classifier} (x_{j, i}),    ~~~ j=\{1, ... , N\}, 
\end{equation}
In this procedure, how to adaptively fuse the dual-modality is the key to successful tracking. In this paper, we predict modality-aware filters $z_v, z_t$ based on each input image pair for dynamic convolutional tracking: 
\begin{equation}
\label{DCTformulation}
[z_v, z_t] = \mathcal{F}_{\theta} ([\mathcal{CNN}(I_v^i); \mathcal{CNN}(I_t^i)]) 
\end{equation}
where $\mathcal{F}_{\theta}$ denotes the proposed MFGNet, $\theta$ is the learnable parameters, [ ; ] is used to denote the concatenate operation, $\mathcal{CNN}$ is the backbone network. With these dynamic filters, we conduct convolution operation on the feature map obtained from backbone. This procedure can be formulated as: 
\begin{equation}
\label{DFNformulation}
\hat{F}_{v} = \mathcal{CNN} (I_v^i)  \otimes z_v; 	~~~   \hat{F}_{t} = \mathcal{CNN} (I_t^i)  \otimes z_t; 
\end{equation}
where the $\otimes$ denotes the convolutional operation. Thanks to the proposed MFGNet, our tracker is more flexible and achieves better tracking results.

\subsection{Overview} 
\textcolor{black}{Different from existing RGB-T trackers which adopt offline learned convolutional kernels for feature extraction only, the key idea of this work is to exploit dynamic kernel prediction for more flexible feature learning. According to the survey \cite{han2021DNNsurvey}, our proposed algorithm belongs to weight prediction which targets directly generating input-adaptive parameters with an independent model at inference time.}  
As shown in Fig. \ref{pipeline}, given the visible and thermal images, we first extract their features with the backbone network. Then, we concatenate the two feature maps and feed them into a dynamic modality-aware filter generation network. Specifically, we design two branches, i.e., the key and query feature generation network, to predict context-aware filters. For each modality, we first conduct matrix multiplication between the key and query features, then, we reshape the output into multiple rectangle kernels. After that, we employ the dynamically generated filters to conduct the convolutional operation on the backbone feature maps. The obtained feature maps are added with backbone feature maps, then, we concatenate the features from the visible and thermal branches as the enhanced multi-modal feature representation. To attain more robust features, we introduce spatial and channel attention to our pipeline before context-aware feature learning. With these enhanced feature maps, we first extract candidate local proposals with Gaussian sampling by following RT-MDNet \cite{jung2018realMDNet}. Inspired by \cite{liu2018picanet, hu2019direction}, we introduce the spatial and temporal RNN into the RGB-T target driven attention model \cite{yang2019learning} to capture global semantic and motion information in a direction-aware manner. The generated attention maps will be used for the mining of global proposals. Then, we train an online classifier for binary classification. The proposal with the maximum score will be chosen as the final tracking result. Similar operations are executed until the end of the testing video sequence.

\subsection{Network Architecture} \label{DILmodule} 

\textcolor{black}{In this subsection, we will discuss the dynamic modality-aware filter generation module, spatial and channel attention, and direction-aware RGB-T target driven attention.}

\subsubsection{Dynamic Modality-Aware Filter Generation Module} \label{dfgnet}

As shown in Fig. \ref{pipeline}, given the visible and thermal videos, we first extract their feature maps with the backbone network. Following the RT-MDNet, we adopt three convolutional layers that are pre-trained on ImageNet dataset \cite{deng2009imagenet} for classification. Specifically, assume the resolution of input images are $W \times H \times 3$ and the output feature maps $F_{v}$ and $F_{t}$ from the backbone are $\frac{W}{8} \times \frac{H}{8} \times 512$. Then, the feature maps from dual modalities will be concatenated together along the channel dimension ($F = [F_{v}; F_{t}]$ whose resolution is $\frac{W}{8} \times \frac{H}{8} \times 1024$) as the input of our dynamic modality-aware filter generation network (i.e., MFGNet).

\textcolor{black}{Although it is an intuitive idea to directly predict the kernel (filter) using fully connected layers, however, this will undoubtedly introduce a large number of learnable parameters. On the other hand, the reshaping from 2D feature maps to vectors may also lose the global context information. In this work, we prefer to utilize a cost-effective way to achieve the modality-aware filter prediction.}  
Our MFGNet contains two parallel and symmetrical branches, i.e., the visible and thermal modality-aware filter generation module. In this section, we will focus on introducing filter generator for visible modality and similar operations are done for the thermal data. For the visible filter generation module, we utilize two independent feature transformation networks, i.e., the $V_q$ and $V_k$ to encode the shared feature maps $F$. In more detail, these transformation networks are implemented with convolutional layers whose kernels are $1 \times 1$. 
\textcolor{black}{We denote the output of $V_k$ and $V_q$ as V-Key-feature $F_v^k$} \textcolor{black}{and V-Query-feature $F_v^q$} \textcolor{black}{(Similarly, we have T-Key-feature $F_t^k$} \textcolor{black}{and T-Query-feature $F_t^q$ for the $T_k$ and $T_q$ network). } 
\textcolor{black}{Specifically, the $V_k$ and $V_q$ transform the concatenated input feature $F$ into $\frac{W}{8} \times \frac{H}{8} \times 512$ ($F_v^k$}\textcolor{black}{) and $\frac{W}{8} \times \frac{H}{8} \times s*s$ ($F_v^q$) , where the $s$ is the size of kernels generated in later procedure.} 
\textcolor{black}{We reshape then transpose the $F_v^k$ into $512 \times W * H$, and conduct matrix multiplication with reshaped $F_v^q \in \mathbb{R}^{W * H \times s*s}$ and get a matrix $\bar{F_v}$ whose dimension is $512 \times s*s$. Similarly, we can have $\bar{F_t}$ for the thermal branch. 
This procedure can be mathematically formulated as:
\begin{equation}
\bar{F_v} = F_v^k * F_v^q, ~~~~~~~~~~ \bar{F_t} = F_t^k * F_t^q. 
\end{equation}
}
We reshape the output into $s \times s \times 512$ and take it as the predicted modality-aware kernels $z_v, z_t$ to carry out dynamic convolutional operation on corresponding backbone feature maps: 
\begin{equation}
\label{dcConv}
\hat{F}_{v} = F_{v} \otimes z_v, ~~~~~~~~~~ \hat{F}_{t} = F_{t} \otimes z_t, 
\end{equation}
where the $\otimes$ denotes dynamic convolutional operation.

The output dynamic modality-aware feature maps are added with backbone feature as enhanced features for each modality (denoted as $F'_{v}$ and $F'_{t}$ respectively): 
\begin{equation}
\label{dcConv}
F'_{v} =  \hat{F}_{v} + F_{v}, ~~~~~~~~~~ F'_{t} = \hat{F}_{t} + F_{t}, 
\end{equation}
Then, the two feature maps are concatenated together as the feature representation of dual modalities, i.e.,  $F' = [F'_{v}; F'_{t}]$. It is worthy to note that our MFGNet can be used together with attention mechanism (such as the popular used spatial and channel attention) to attain a more robust representation as described in the following subsection.

\begin{figure}
\center
\includegraphics[width=3in]{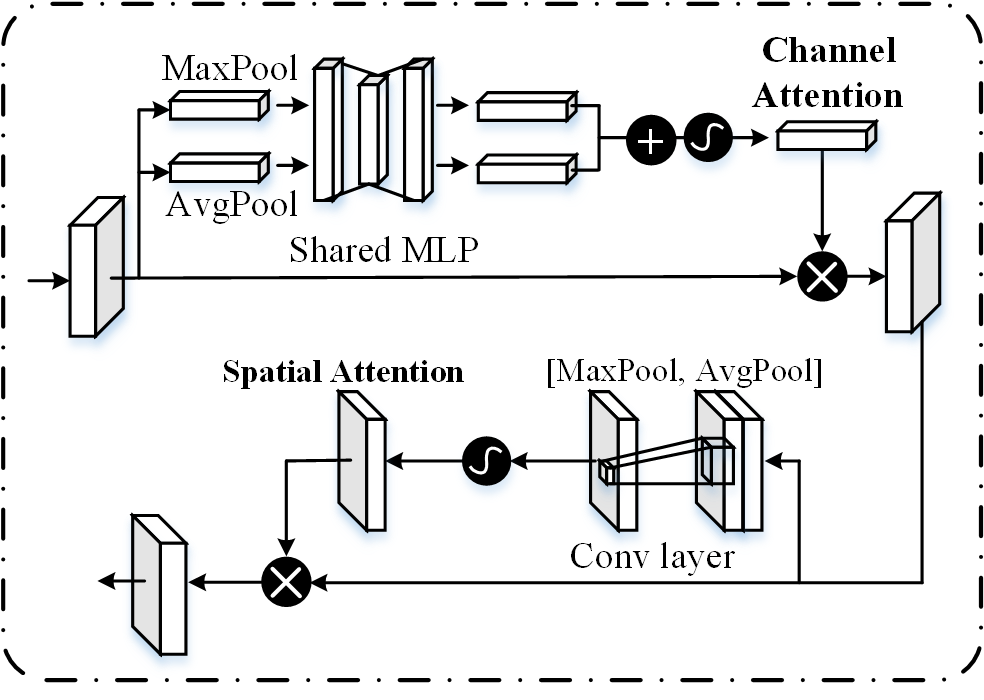}
\caption{The illustration of the used spatial and channel attention network for more robust feature learning.}  
\label{SCNet}
\end{figure}

\subsubsection{Spatial and Channel Attention} \label{SCNet}  
In this paper, the Convolutional Block Attention Module (CBAM) \cite{woo2018cbam} is used to emphasize meaningful features along channel and spatial axes to learn \emph{what} and \emph{where} to attend. As shown in Fig. \ref{SCNet}, CBAM first attends the input feature with channel attention, then, it attends the channel-refined feature from the perspective of spatial attention. Specifically, we use max-pooling and average pooling operations to attain two feature vectors. Then, these features will be fed into a shared multi-layer perceptron module and the output features will be summarized and passed to an activation layer. Therefore, the channel-refined feature can be obtained by multiplying the channel attention and original input feature.

After that, we employ max-pooling and average pooling to get two feature maps. We adopt a convolutional layer to fuse these features and input them into an activation layer to obtain spatial attention. Finally, the refined features can be obtained by multiplying the spatial attention and input feature maps. Our experimental results demonstrate that the tracking performance can be further improved with such an attention scheme. More details will be introduced in the ablation study in section \ref{ablationStudy}.

\subsubsection{Direction-Aware RGB-T Target driven Attention} \label{daTANet} 
With the enhanced RGB-T features, we can directly train an online classifier for tracking as many previous RGB-T trackers do. Most of them adopt local search for tracking which makes their tracker sensitive to challenging factors as noted in \cite{yang2019learning}. Yang et al. \cite{yang2019learning} propose to conduct joint local and global searches with a target-aware attention network. The spatial information is considered in their model, while the temporal information that is also very important for target localization is ignored.

\begin{figure} 
\center
\includegraphics[width=3.5in]{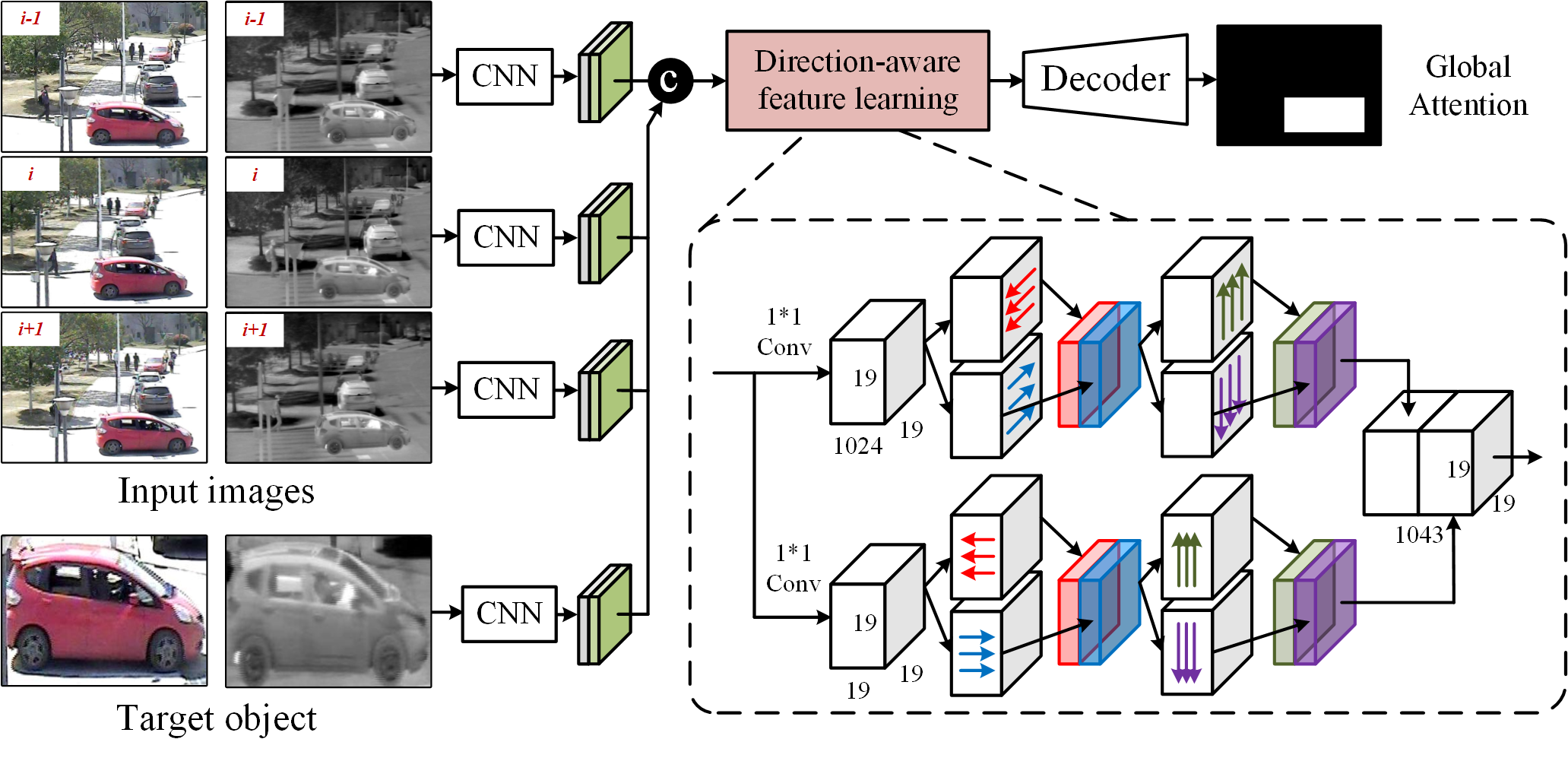}
\caption{Illustration of our proposed direction-aware target driven attention network (daTANet) for global search. }
\label{RGBT_AttentionModule}
\end{figure}

In this work, we introduce the spatial and temporal RNN to capture the global semantic and motion feature in a \emph{direction-aware} manner. As shown in Fig. \ref{RGBT_AttentionModule}, given continuous video frame pairs and target templates initialized in the first frame (these images are all resized into $300 \times 300$), we first extract their features with a deep neural network (ResNet-18 \cite{he2016deep} is adopted in this paper). We concatenate the feature maps from two different layers to better capture the low-level fine-grained information and high-level semantic information. Then, the visible and thermal feature maps of the same image pair are summarized and features from different video frames are concatenated along with the channel dimension. Therefore, we can get a feature map whose dimension is $3072 \times 19 \times 19$. To better capture the target object from the spatial and temporal view, we introduce the LSTM network \cite{hochreiter1997lstm} to ``sweep" the encoded feature maps from various directions to capture the \emph{direction-aware features}. Assume we have activation function $f$, input gate $i_i$, output gate $o_t$ and forget gate $f_t$, therefore, the key formulation of LSTM can be written as: 
\begin{flalign}
&~~~~\hat{x_t} = \textbf{W} {x_t}& \\
&~~~~f'_t = \delta (\textbf{W}_f x_t + b_f)& \\
&~~~~r_t = \delta (\textbf{W}_r x_t + b_r) \\
&~~~~c_t = f_t \odot c_{t-1} + (1-f_t) \odot \hat{x_t} \\
&~~~~h_t = r_t \odot g(c_t) + (1-r_t) \odot x_t
\end{flalign} 
where $\odot$ is the Hadamard product, i.e., the element-wise multiplication. $x_t, c_t$ and $h_t$ represent the input features, cell states and cell outputs, respectively. $\textbf{W}$ and $b$ are learnable parameters. In this work, we first use a $1 \times 1$ convolutional operator to transform the input features into $1024 \times 19 \times 19$. We utilize the spatial LSTM to obtain direction-aware features from four directions, as shown in Fig. \ref{RGBT_AttentionModule}. More detail, a bi-directional LSTM is adopted to sweep along the row and column of feature maps sequentially. And the output feature maps are concatenated along with channel dimension. For the temporal view, we first encode the input features into $19 \times 19 \times 19$, then, we transpose the first and second dimension and use another bi-directional LSTM to sweep for the motion feature. We transpose the attained feature maps back and take it as the motion features, therefore, we can get the final direction-aware feature maps whose dimension is $1043 \times 19 \times 19$ by concatenating the spatial and temporal features. Following \cite{yang2019learning}, we also use a decoder network to upsample the resolution of obtained RGB-T target-driven attention maps. More detail, the decoder network is composed of five groups of upsample module, and each upsample module is consisted of three transposed convolutional layers and one upsample layer. More details can be found from our source code. The ground truth mask used for the training of this network is also obtained by whiting the target object regions and blacking the background regions, more details can be found in \cite{yang2019learning}.

\subsection{Training and Tracking} 
Given the RGB $I_v^i$ and thermal image $I_t^i$ pair, some positive and negative samples are extracted using Gaussian sampling around the ground truth bounding box. Then, we use three convolutional layers to extract their features respectively. Specifically, the features from Conv3 are fed into the attention based dynamic convolutional module to acquire the enhanced features. Following RT-MDNet \cite{jung2018realMDNet}, we also introduce multi-domain layers into our model in the training phase. The cross-entropy loss $\mathcal{L}_{cls}$ and instance embedding loss function $\mathcal{L}_{inst}$ are adopted for the optimization: 
\begin{equation}
\label{LclsFunction}
\mathcal{L} = \mathcal{L}_{cls} + \alpha \cdot \mathcal{L}_{inst}, 
\end{equation}
where $\alpha$ is a hyper-parameter that controls the balance between $\mathcal{L}_{cls}$ and $\mathcal{L}_{inst}$. Following RT-MDNet, we set the $\alpha$ as 0.1 in all our experiments in this paper. Specifically speaking, the $\mathcal{L}_{cls}$ can be formulated as: 
\begin{equation}
\label{LclsFunction}
\mathcal{L}_{cls} = -\frac{1}{N} \sum_{j=1}^{N}\sum_{c=1}^{2} [\textbf{y}_j]_{c\hat{d}(m)} \cdot log([\sigma_{cls} (\textbf{f}_j^{\hat{d}(m)})]_{c\hat{d}(m)})
\end{equation}
where $\textbf{y}_j$ is the binary ground truth label; the element of $[\textbf{y}_j]_{cd}$ will be 1 if the class of a bounding box in the domain $d$ is $c$, otherwise its element will be 0. $m$ is the index of iteration. $\textbf{f}_j$ is the predicted score of $j$-th proposal by online learned classifier. $\sigma_{cls}(*)$ and $\sigma_{inst}(*)$ are softmax functions to normalize the predicted scores.  
The formulation of $\mathcal{L}_{inst}$ can be written as: 
\begin{equation}
\label{LinstFunction}
\mathcal{L}_{inst} = -\frac{1}{N} \sum_{j=1}^{N}\sum_{d=1}^{D} [\textbf{y}_j]_{+d} \cdot log([\sigma_{inst} (\textbf{f}_j^{d}]_{+d}) 
\end{equation}
where $D$ is the number of domains (i.e., videos) in the training dataset. 
As noted in RT-MDNet, the instance embedding loss is only applied to positive examples using the positive channel in Eq. \ref{LinstFunction} (denoted by +). More details about the two loss functions can be found in the RT-MDNet \cite{jung2018realMDNet}.

In the tracking phase, we first extract some proposals around the previous tracking result using the Gaussian sampling method and feed them into the tracking network to get the final classification scores. The proposal with the highest score is regarded as the tracking result of the current frame. After that, the bounding box regression model trained on the first frame is used to refine the predicted location of the target object. In addition, we also adopt the short-term and long-term update strategy and hard mini-batch mining which is also used in \cite{Nam2015Learning, jung2018realMDNet} to achieve more robust tracking. In this procedure, we will switch to the global search (i.e., search the target object around the global attentive regions) if the classification score is lower than 0 for continuous $\mathcal{N}$ frames.

\begin{table*} 
\center
\scriptsize  
\caption{Tracking results (PR/SR) on the RGBT-210 and RGBT-234 dataset.} \label{benchmarkResults}
\begin{tabular}{c|cccccccccccccc}
\hline \toprule [0.7 pt]
\textbf{Algorithm}   	&SiamFC \cite{Bertinetto2016SiameseFC} 	&Staple \cite{bertinetto2016staple} 	&BACF \cite{kiani2017learning} 	&SRDCF \cite{lukezic2017CSRDCF} 	&SGT \cite{li2017weightedRGBT210} 	&ECO \cite{danelljan2017eco} 	&DiMP \cite{bhat2019DiMP} 	&mfDiMP \cite{zhang2019mfDiMP} 	&RT-MDNet \cite{jung2018realMDNet} 	&Ours \\
\hline 
\textbf{RGBT-210}   &0.586/0.412 	&0.595/0.429 	&0.616/0.451 	&0.619/0.442 	&0.675/0.430 	&0.690/0.498 	&0.719/0.513 	&0.786/0.555 	&0.715/0.467	&0.749/0.494 	 \\
\hline \toprule [0.7 pt]
\textbf{Algorithm}    &JSR \cite{liu2012fusionJSR} 	&KCF \cite{Henriques2015High} 	&CFNet \cite{valmadre2017CFNet} 	&SiamDW \cite{zhipeng2019siamDW} 	&MDNet \cite{Nam2015Learning} 	&DPANet \cite{zhu2019DPANet} 	&SGT \cite{li2017weightedRGBT210} 	&SOWP \cite{kim2015sowp} 	&RT-MDNet  \cite{jung2018realMDNet} 	&Ours \\
\hline 
\textbf{RGBT-234}   &0.343/0.234  	&0.463/0.305 	&0.551/0.390  	&0.604/0.397  	&0.722/0.495  	&0.766/0.537  	&0.720/0.472 	&0.696/0.451  	&0.758/0.515  	&0.783/0.535 	 \\
\hline \toprule [0.7 pt]
\end{tabular}
\end{table*}

\section{Experiment} 

\subsection{Dataset and Evaluation Metric} \label{datasetMetric}
For the training of our tracker, we adopt the GTOT-50 dataset which is the first standard benchmark for RGB-T tracking proposed by Li et al. in \cite{Li2016Learning}. It consists of 50 RGB and thermal video pairs with ground truth, which are collected under different scenarios and conditions. In addition, the videos are annotated with attributes and can be divided into seven subsets according to the attributes for evaluating the sensitivity of the algorithm to these attributes. For the testing, two large-scale RGB-T tracking benchmark datasets are used in the experiments, including RGBT-210 and RGBT-234 datasets. RGBT-210 dataset is proposed in \cite{li2017weightedRGBT210} which contains 210 video sequences and RGBT-234 is another large-scale RGB-T tracking dataset proposed in \cite{li2019rgb}. It is worthy to note that RGBT-234 is extended from RGBT-210 dataset \cite{li2017weightedRGBT210} and contains 234 RGB and thermal video pairs. The total frame number of RGBT-234 is about 234,000, and the longest sequence has 8,000 frames. In addition, this dataset is annotated with 12 kinds of attributes. For the evaluation of GTOT-50 dataset, we train our model on the RGBT-234 dataset. 

In our experiments, the precision rate (PR) and success rate (SR) are used as the evaluation criteria to compare the tracking results. PR is the percentage of frames whose output position is within the ground truth threshold distance, SR is the percentage of frames whose overlap rate between the output bounding box and the ground truth boundary box is greater than the threshold value. We use the area under the curves of the success rate as the representative SR for quantitative performance evaluation. More specifically, the threshold of distance is set to 20 pixels for RGBT-210 and RGBT-234 datasets respectively, and the threshold for overlap is set to 0.6 for both datasets. The threshold of distance is set to 5 pixels for GTOT-50 dataset due to many targets are small objects in this dataset.

\subsection{Implementation Details}  
For the training of RT-MDNet\footnote{\textcolor{magenta}{\url{https://github.com/BossBobxuan/RT-MDNet}}} based RGB-T tracker, the number of positive and negative samples are set as 32 and 96 respectively. The extracted sample is regarded as positive sample if the IoU between it and the ground truth is in the range 0.7 to 1, and it is regarded as negative sample if the IoU is less than 0.5. All of the samples are resized to $107 \times 107 \times 3$. In the tracking phase, we update the model online every 10 frames, or the tracking failure is detected. The initial learning rate is 0.0001, the decay is 0.0005 and the momentum is 0.9. We train 1k epochs to search for a better model for tracking.

For the training of our RGB-T target driven attention network, we randomly select 300 video sequences from GOT-10k dataset \cite{huang2019got}. We transform the color images into gray images and take them as the corresponding thermal image to train our attention network for 30 epochs. The batch size is 5, the initial learning rate is 1e-5, and the Adagrad \cite{duchi2011adaptive} is adopted for the optimization of our network. The experiments are implemented based on Python 3.7, PyTorch 1.0.1 \cite{paszke2019pytorch} on a machine with CPU I7-6700k and four GPUs Tesla P100.

\subsection{Comparison on Public RGB-T Benchmarks} \label{comparedTrackers}
In this section, we compare our proposed algorithm with other trackers on three large-scale RGB-T tracking benchmark datasets, including SiamFC \cite{Bertinetto2016SiameseFC}, Staple \cite{bertinetto2016staple}, BACF \cite{kiani2017learning}, SRDCF \cite{lukezic2017CSRDCF}, SGT \cite{li2017weightedRGBT210}, ECO \cite{danelljan2017eco}, DiMP \cite{bhat2019DiMP}, mfDiMP \cite{zhang2019mfDiMP}, RT-MDNet \cite{jung2018realMDNet}, JSR \cite{liu2012fusionJSR}, KCF \cite{Henriques2015High}, CFNet \cite{valmadre2017CFNet}, SiamDW \cite{zhipeng2019siamDW}, MDNet \cite{Nam2015Learning}, DPANet \cite{zhu2019DPANet}, SOWP \cite{kim2015sowp}, C-COT \cite{danelljan2016CCOT}, DAFNet \cite{gao2019DAFNet}, FANet \cite{zhu2018fanet} and MANet \cite{long2019MANet}.

\textbf{Results on RGBT-210 Dataset.} 
As shown in Table \ref{benchmarkResults}, our tracker achieves 0.749/0.494 on the RGBT-210 dataset based on the PR/SR, while the baseline method RT-MDNet attains 0.715/0.467. It is easy to find that our tracker improves the baseline tracker significantly on both evaluation criteria, and this experiment fully validated the effectiveness of our proposed modules for RGB-T tracking. We can also find that Siamese network based tracker like SiamFC (0.586/0.412) and correlation filter based trackers such as Staple (0.595/0.429) and SRDCF (0.619/0.442) are all worse than our tracking results. This fully demonstrates the advantages of our tracker compared with these trackers. In addition, our tracker also achieves better tracking results compared with ECO and DiMP on the PR, which are all popular and strong trackers proposed in recent years. Our tracker also attains comparable performance with ECO on the SR. These experiments all show that our proposed modules can capture more discriminative features for RGB-T tracking. Our tracker is worse than DiMP on the SR criterion, we think this may be caused by: 1). DiMP adopts a better proposal generation module IoUNet, while we only use a simple Gaussian sampling operation; 2). DiMP adopts a much deeper network and hierarchical feature maps for feature representation, while we only use three convolutional layers for the feature encoding. We will consider improving our tracker from these two aspects of our future works.

\textbf{Results on RGBT-234 Dataset.}
As shown in Table \ref{benchmarkResults}, we can find that the baseline tracker RT-MDNet achieves 0.758/0.515 on the RGBT-234 dataset which is significantly better than most of the compared trackers, such as JSR (0.343/0.234), KCF (0.463/0.305) and MDNet (0.722/0.495). Compared with this baseline, our tracker can improve it to 0.783/0.535 with the help of our proposed module. This experiment also fully validates the effectiveness of our proposed network for RGB-T tracking.

\begin{figure}
\center
\includegraphics[width=3.5in]{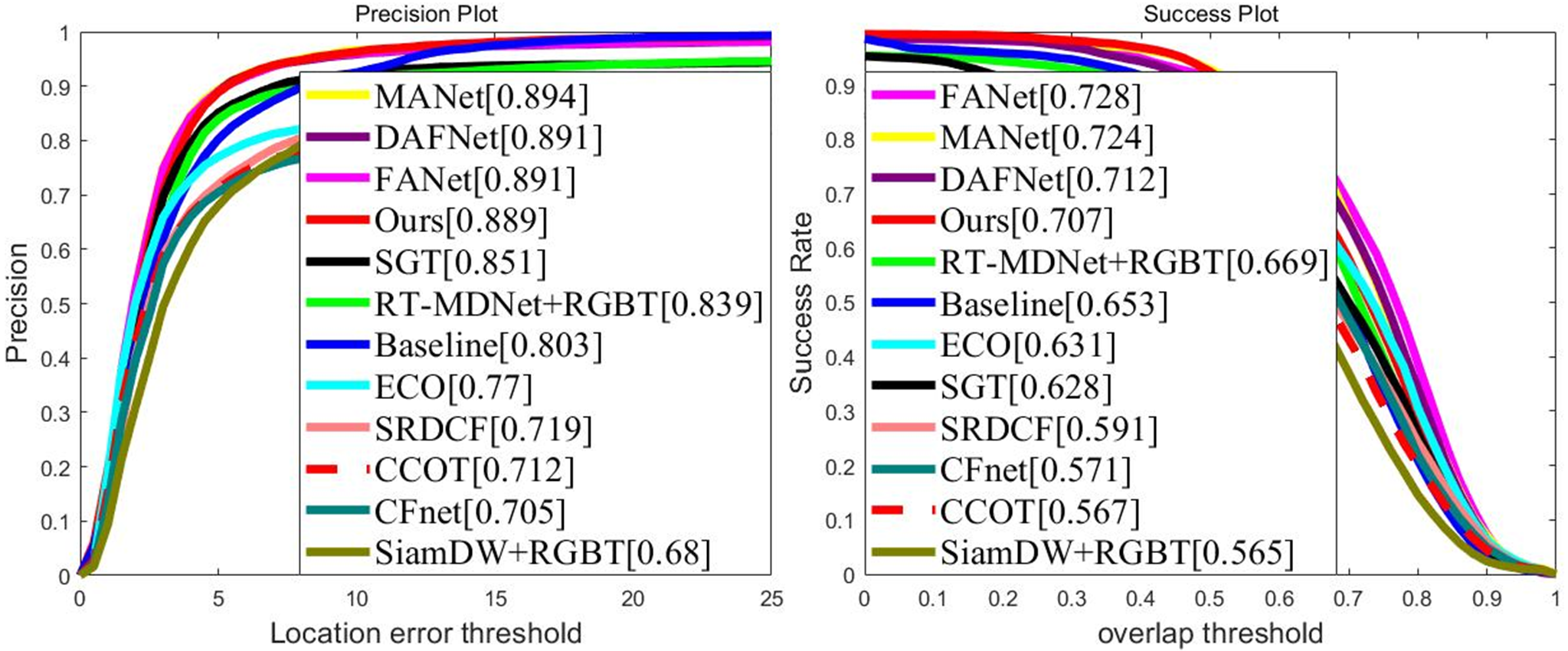}
\caption{Tracking results on the GTOT-50 dataset. }
\label{GTOT50_results}
\end{figure}

\textbf{Results on GTOT-50 Dataset.}
As shown in Fig. \ref{GTOT50_results}, our baseline approach achieves 0.803/0.653 on the GTOT-50 dataset on the PR/SR respectively. Our tracker can attain 0.889/0.707 on this benchmark which is significantly better than the baseline approach. This also fully validates the effectiveness of our proposed modules for RGB-T tracking. Compared with other RGB-T trackers, our tracker is also better than SGT (0.851/0.628), ECO (0.770/0.631), SRDCF (0.719/0.591), CFNet (0.705/0.571) and SiamDW (0.680/0.565). This also demonstrates the advantages of our tracker. Our tracker is slightly worse than FANet (0.891/0.728), DAFNet (0.891/0.712), and MANet (0.894/0.724) which were all developed based on MDNet. They sacrifice the running efficiency (less than 1 FPS) for the final tracking performance which makes their trackers impossible for practical application.

\subsection{Ablation Study} \label{ablationStudy}
To analyze the effectiveness of each component in our algorithm, we conduct extensive ablation studies on the RGBT-210 dataset.

\begin{table*} 
\center
\caption{Component analysis on the RGBT-210 dataset.} \label{CMAnalysis}
\begin{tabular}{c|cccccc|c}
\hline \toprule [1 pt]
&Baseline   &+nDFNet  &+DFNet  &+SCNet	 &+nTANet  &+daTANet	&PR/SR			 \\
\hline 
Tracker-1 	&\checkmark &  	&    		  &    	&   		&   																&0.715/0.467   				\\
Tracker-2  	&\checkmark &\checkmark  &       	&   	&    			&   												&0.714/0.470 			   		\\
Tracker-3 	&\checkmark &    	&\checkmark   &    	&   			&   												&0.736/0.487   			   		\\
Tracker-4 	&\checkmark &    &\checkmark   	&\checkmark    	&   		&   								&0.744/0.487   			   		\\
Tracker-5 	&\checkmark  &    	&\checkmark  &\checkmark    	&\checkmark   			&   		&0.749/0.494   			    		 \\
Ours 			&\checkmark  &    	&\checkmark  &\checkmark    	&   			&\checkmark   		&0.759/0.495   			    		 \\
\hline \toprule [1 pt]
\end{tabular}
\end{table*}

\subsubsection{Component Analysis} 
To check how much of each component in our model contributes to the final tracking performance, we design the following component analysis:   

$\bullet$ Baseline: the RGB-T version of RT-MDNet; 

$\bullet$ +nDFNet: a naive version of the dynamic filter network. Specifically, we directly concatenate the visible and thermal features and only predict one set of filters for later dynamic convolutional operation; 

$\bullet$ +DFNet: we integrate the dynamic modality-aware filter network into the RT-MDNet; 

$\bullet$ +SCNet: we integrate the spatial and channel attention network into our baseline; 

$\bullet$ +nTANet: we integrate the naive version RGB-T target driven attention network for a global search for the RGB-T tracking; 

$\bullet$ +daTANet: we integrate the direction-aware RGB-T TANet for global search; 

$\bullet$ Ours: we utilize all the aforementioned modules, i.e., the Baseline + DFNet + SCNet + daTANet, for RGB-T tracking.

As shown in Table \ref{CMAnalysis}, the baseline method achieves 0.715/0.467 on the PR/SR on the RGBT-210 dataset, respectively. When integrated with the DFNet proposed in this paper, the results can be improved to 0.736/0.487. This experiment fully validated the effectiveness of our designed modality-aware filter generation module. We also find that the baseline + nDFNet only achieves 0.714/0.470 on the RGBT-210 dataset which is slightly better than the baseline method on the SR. Our modality-aware dynamic filter network attains better results than nDFNet, which fully validated the importance of modality-aware filter generation. In addition, we also check the effectiveness of spatial and channel mechanisms, i.e., the SCNet, as shown in Table \ref{CMAnalysis}. It is easy to find that the PR can be improved from 0.736 to 0.744, while the SR is comparable by utilizing the SCNet (i.e., the Tracker-4 \emph{vs.} Tracker-3). This experiment demonstrates that the SCNet also contributes to our RGB-T tracker.

The aforementioned RGB-T trackers adopt the local search for tracking which is easily influenced by challenging factors. When integrating Tracker-4 with the nTANet, the tracking performance can be improved from 0.744/0.487 to 0.749/0.494. These results show the benefits of global search supported by nTANet. For the comparison between Tracker-5 and Ours (the only difference is that Tracker-5 utilizes the nTANet, while Ours uses the daTANet), we can find that Ours can improve the PR/SR from 0.749/0.494 to 0.759/0.495. These experimental results demonstrate the advantages and effectiveness of the RNN which can capture spatial and temporal information for accurate global attention prediction. Finally, we can observe that all the aforementioned experimental results and comparisons demonstrate that our proposed modules are beneficial for RGB-T tracking.

\subsubsection{Analysis on Size of Kernels}  
\begin{figure} 
\center
\includegraphics[width=3.5in]{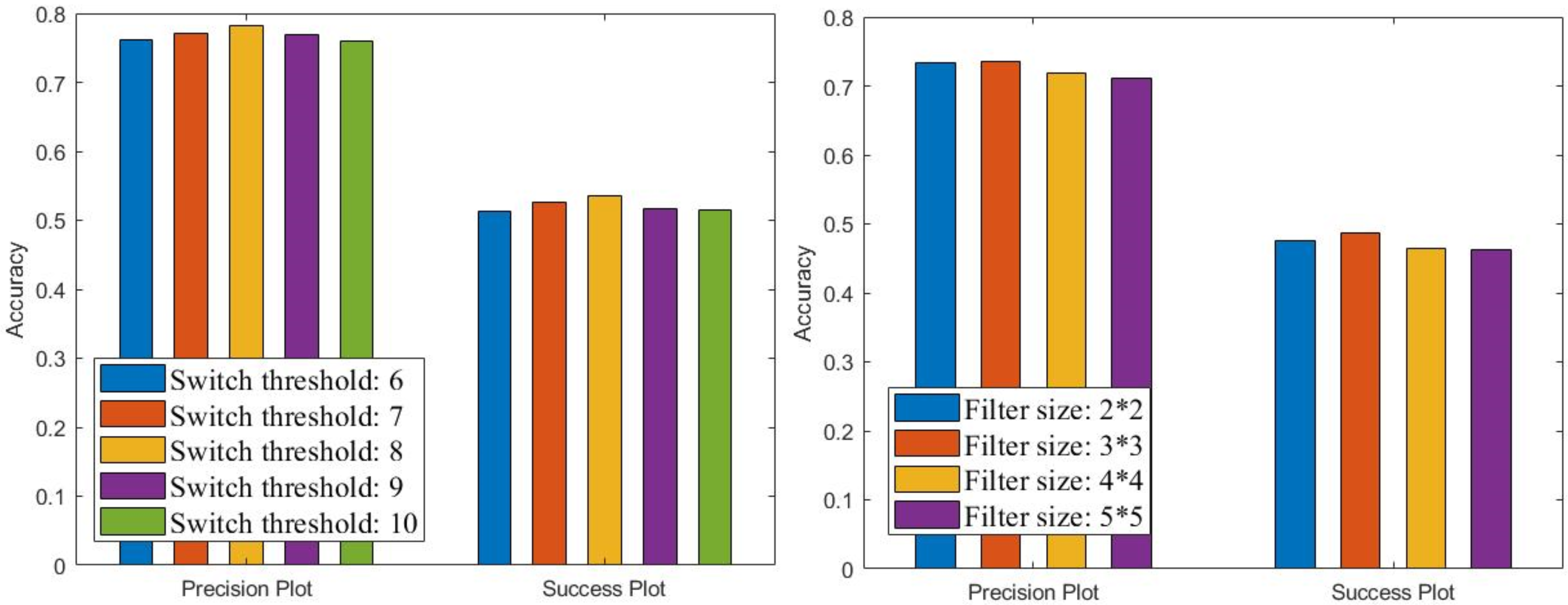}
\caption{ \textcolor{black}{Tracking results with different threshold parameters $\mathcal{N}$ on RGBT-234 dataset (left sub-figure), and various kernel size on RGBT-210 dataset (right sub-figure).} }  
\label{kernelSizethresholdAnalysis}
\end{figure}

As shown in Fig. \ref{kernelSizethresholdAnalysis} (right sub-figure), we also report the tracking results with various kinds of convolutional filters on the RGBT-210 dataset. It is easy to find that we can get the best performance when the kernel size is 3. We also find that the tracking performance becomes worse when the kernel size is larger than 4. When we set it as 5, the tracking results are even worse than the baseline approach. Therefore, we predict the dynamic convolutional filters whose dimension is $3 \times 3$ for all our experiments.

\subsubsection{Analysis on Threshold Parameters}   

In this work, when to utilize the global or local search (i.e., use direction-aware TANet or not) plays an important role in RGB-T tracking. Because the global attention prediction is not always accurate it may easily lead to model drift. In this paper, we simply select the search strategy according to the response score of previous tracking results by following \cite{yang2019learning}. Specifically speaking, we monitor the times of \emph{tracking failure} (the RT-MDNet regards the tracking as failed if the response score is lower than $0$) and denote it as $\mathcal{N}$. With a smaller value of $\mathcal{N}$, we can make the tracker focus more on the global search, while a larger $\mathcal{N}$ will make our tracker more ``conservative" (i.e., tend to use proposals from local search window). As shown in Fig. \ref{kernelSizethresholdAnalysis} (left sub-figure), we report the tracking results under different threshold parameters $\mathcal{N}$ on the RGBT-234 dataset. It is easy to find that when the threshold parameter is set as 8, we can get the best tracking performance compared with others. Our method is simple yet effective and improves the tracking results significantly, as shown in Table \ref{CMAnalysis}. Therefore, we can conclude that our proposed direction-aware TANet can improve the robustness of the baseline tracker significantly.

\begin{figure*} 
\center
\includegraphics[width=7in]{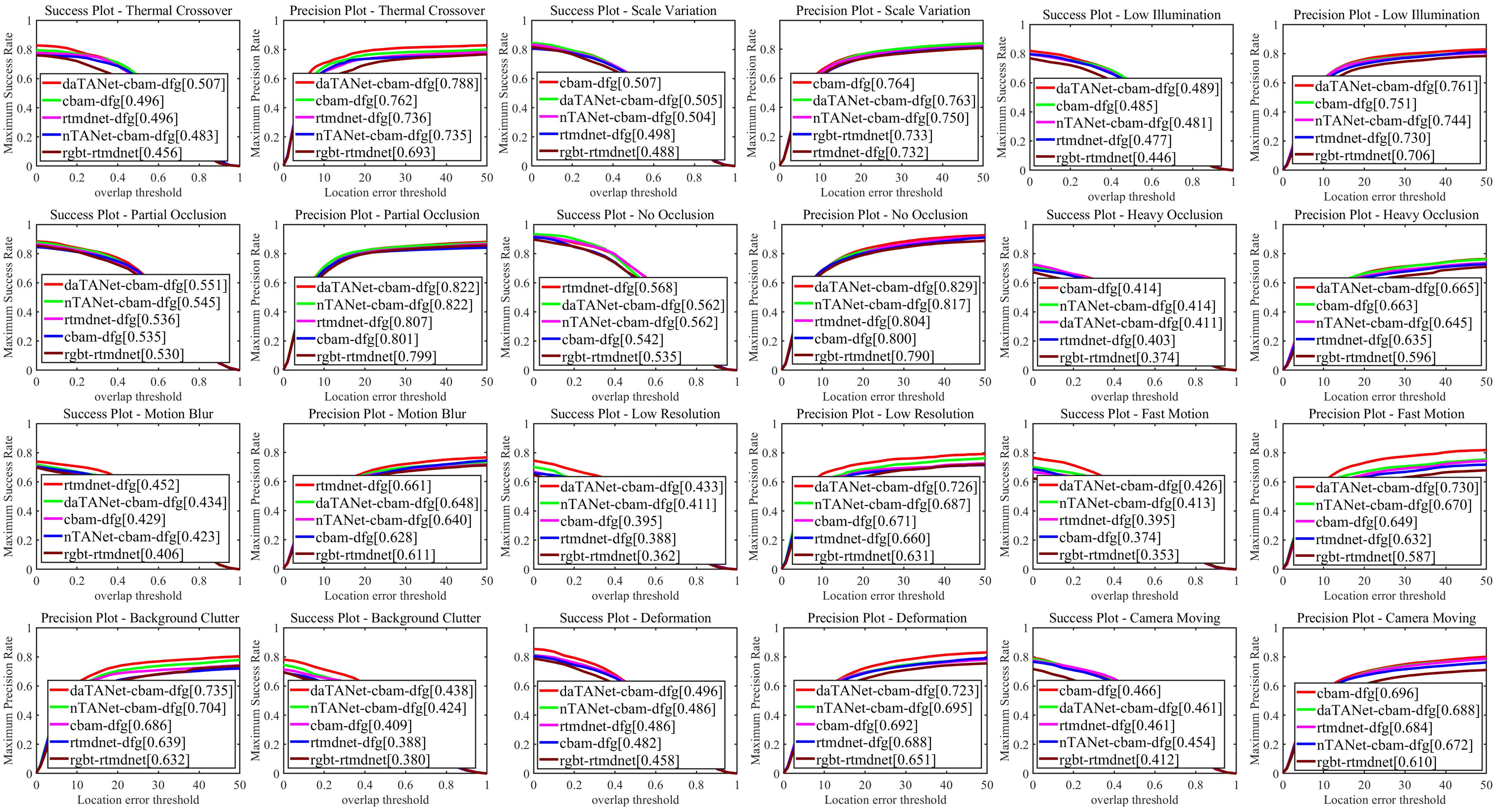}
\caption{Tracking results under each attribute on the RGBT-210 dataset (rgbt-rtmdnet is our baseline method; cbam-dfg denotes the baseline tracker with the SCNet and DFG; rtmdnet-dfg denotes tracker with DFG only; nTANet-cbam-dfg is the baseline tracker with the SCNet, DFG and nTANet; daTANet-cbam-dfg is our final tracker).}  
\label{attribute210Results}
\end{figure*}

\subsubsection{Attribute Analysis} 
To validate the effectiveness of our proposed modules for handling challenging factors. We report the tracking results under each attribute on the RGBT-210 and RGBT-234 datasets. As shown in Fig. \ref{attribute210Results}, we can find that our proposed modules improve the baseline method on all the mentioned challenging factors in the RGBT-210 dataset. For example, the baseline tracker RT-MDNet achieves 0.799/0.530 on the challenge of \emph{partial occlusion}, while our tracker attains 0.822/0.551 which is significantly better than the baseline method. Besides, we can find that our final tracker, i.e., daTANet-cbam-dfg, can achieve better performance than other versions on most of the challenging factors, including \emph{thermal crossover, background clutter, deformation, fast motion, heavy occlusion, low illumination} and \emph{low resolution}. This also fully validated the advantages of our proposed modules.

In addition, we also compare our tracker with other RGB-T trackers like SGT and MANet on each attribute of RGBT-234 dataset, we can find that our tracker achieves better results. Our performance is comparable with trackers like FANet and MANet in Table \ref{attribute234Result}. Therefore, we can conclude that our proposed modules significantly improve the baseline tracker RT-MDNet, and our tracker can also be competitive with existing state-of-the-art trackers.

\begin{table*} 
\center
\scriptsize 
\caption{\textcolor{black}{Tracking results (PR/SR) under each attribute on RGBT-234 dataset.} } \label{attribute234Result}
\begin{tabular}{c|cccccccccccccccc}
\hline \toprule [1 pt]
\textbf{Attribute}   &\textbf{SOWP} \cite{kim2015sowp}  &\textbf{CFNet} \cite{valmadre2017CFNet} &\textbf{C-COT} \cite{danelljan2016CCOT} &\textbf{SGT} \cite{li2017weightedRGBT210} &\textbf{DAFNet} \cite{gao2019DAFNet} &\textbf{FANet} \cite{zhu2018fanet} &\textbf{MANet} \cite{long2019MANet}   &\textbf{RT-MDNet} \cite{jung2018realMDNet}     &\textbf{Ours} \\ 
\hline 
{Low Resolution} 	&0.725/0.462      &0.551/0.365            &0.731/0.494  &0.751/0.476  &{{0.818}}/{{0.538}}    &{{0.795}}/{{0.532}} &0.757/0.515   &0.762/0.478    &0.793/0.495         \\
{Motion Blur}         &0.639/0.421      &0.357/0.271      &0.673/0.495  &0.647/0.436 &{{0.708}}/0.500 &{{0.700/0.503}} &{{0.726}}/{{0.516}}  &0.671/0.475   &0.737/0.510         \\
{No Occlusion}          &0.868/0.537      &0.764/0.563&{{0.888}}/{{0.656}}  &0.877/0.555 &{{0.900}}/0.636   &0.893/{{0.657}} &0.887/{{0.646}}  &0.887/0.627   &0.920/0.640         \\
{Heavy Occlusion}      &0.570/0.379      &0.417/0.290      &0.609/0.427  &0.592/0.394 &{{0.686}}/{{0.459}} &0.665/0.458 &{{0.689}}/{{0.465}}  &0.645/0.429   &0.662/0.443         \\
{Partial Occlusion}      &0.747/0.484      &0.597/0.417      &0.741/0.541 &0.779/0.513 &{{0.859}}/{{0.588}} &{{0.866}}/{{0.602}} &0.816/0.566  &0.814/0.551 &0.843/0.580         \\
{Scale Variation}       &0.664/0.404      &0.598/0.433      &0.762/{{0.562}} &0.692/0.434 &{{0.791}}/0.544 &{{0.785}}/{{0.563}} &0.777/0.542  &0.754/0.522 &0.761/0.528         \\
{Thermal Crossover} &0.701/0.442      &0.457/0.327      &{{0.840}}/{{0.610}} &0.760/0.470 &{{0.811}}/{{0.583}} &{{0.766}}/0.549 &0.754/0.543   &0.793/0.557 &0.818/0.558         \\
{Low Illumination}      &0.723/0.468      &0.523/0.369      &0.648/0.454 &0.705/0.462 &{{0.812}}/{{0.542}} &{{0.803}}/{{0.548}} &0.769/0.513 &0.743/0.505 &0.791/0.542         \\
{Fast Motion}         &0.637/0.387      &0.376/0.250      &0.628/0.418 &0.677/0.402 &{{0.740}}/{{0.465}} &0.681/0.436 &{{0.694}}/{{0.449}}  &0.650/0.411 &0.725/0.446         \\
{Deformation}         &0.650/0.460      &0.523/0.367      &0.634/0.463 &0.685/0.474 &{{0.741}}/0.515 &{{0.722}}/{{0.526}} &{{0.720}}/{{0.524}}   &0.671/0.472 &0.721/0.508         \\
{Camera Moving}         &0.652/0.430      &0.417/0.318       &0.659/0.473 &0.667/0.452 &{{0.723}}/{{0.506}} &{{0.724}}/{{0.523}} &{{0.719}}/{{0.508}}  &0.677/0.467 &0.732/0.504         \\
{Background Clutter}      &0.647/0.419      &0.463/0.308      &0.591/0.399 &0.658/0.418 &{{0.791}}/{{0.493}} &{{0.757}}/{{0.502}} &0.739/0.486 	&0.690/0.442 &0.743/0.459         \\
\hline \toprule [1 pt]
\end{tabular}
\end{table*}

\subsubsection{Efficiency Analysis} 
Our tracker can run at 3.37 FPS on a video whose resolution is $630 \times 460$, including the time spend on the baseline local search tracker (RT-MDNet + SCNet + DFG) and global attention prediction (daTANet).  Actually, our tracker is faster than 3.37 FPS, if we only use daTANet when tracking failure is detected. It is worthy to note that our tracker can run at 10.75 FPS without the daTANet.

\begin{figure*} 
\center
\includegraphics[width=7in]{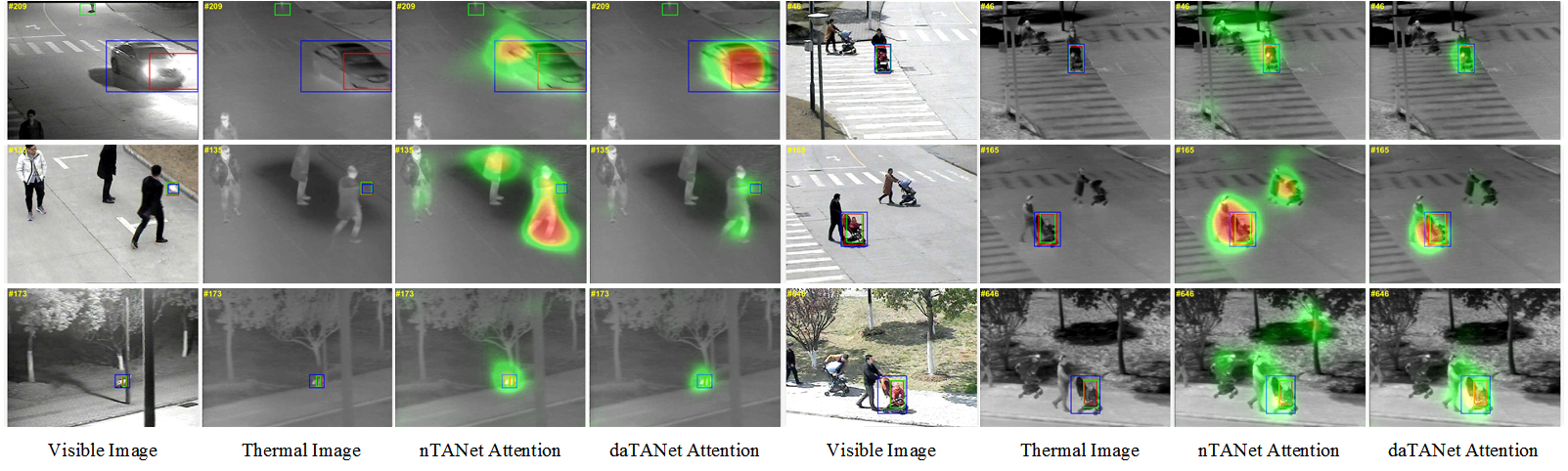}
\caption{Visualization of attention predicted by nTANet and daTANet. }
\label{TANetattentionVIS}
\end{figure*}

\subsection{Visualization} 

In this section, we give some visualization of learned global target-aware attention maps, feature maps, and tracking results to better understand the effects of each component in our model. 

\textbf{Feature Map \& Global Attention Map:} 
As shown in Fig. \ref{TANetattentionVIS}, we can find that the global attention predicted by our nTANet and daTANet can focus on the real target object. Therefore, we can improve the baseline approach which only adopts a local search for RGB-T tracking significantly. It is also easy to find that the attention predicted by daTANet is more accurate than nTANet. For example, the main body of \emph{car} is highlighted by our daTANet while the nTANet only focuses on its tail. The daTANet pays more on the real \emph{baby carriage}, while the nTANet can be distracted by another similar one. This fully validated the effectiveness and advantages of our spatial-temporal RNN for RGB-T tracking.

\begin{figure} 
\center
\includegraphics[width=3.3in]{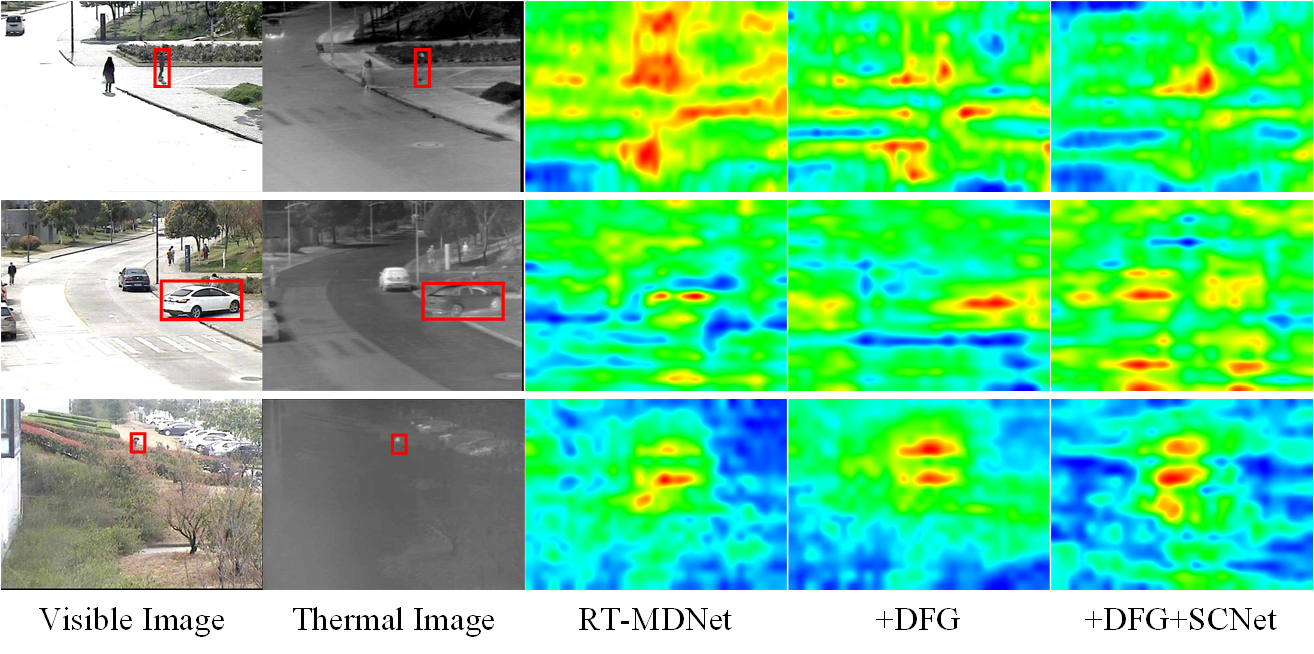}
\caption{Visualization of feature maps of our model. }
\label{featMAPVIS}
\end{figure}

In Fig. \ref{featMAPVIS}, we give some visualization of the learned convolutional feature maps. We can find that the feature maps learned by our backbone network (+DFG, +DFG+SCNet) have a better response to the target object than the baseline approach RT-MDNet. These results all demonstrate the advantages of our proposed modules for RGB-T tracking.

\begin{figure*} 
\center
\includegraphics[width=7in]{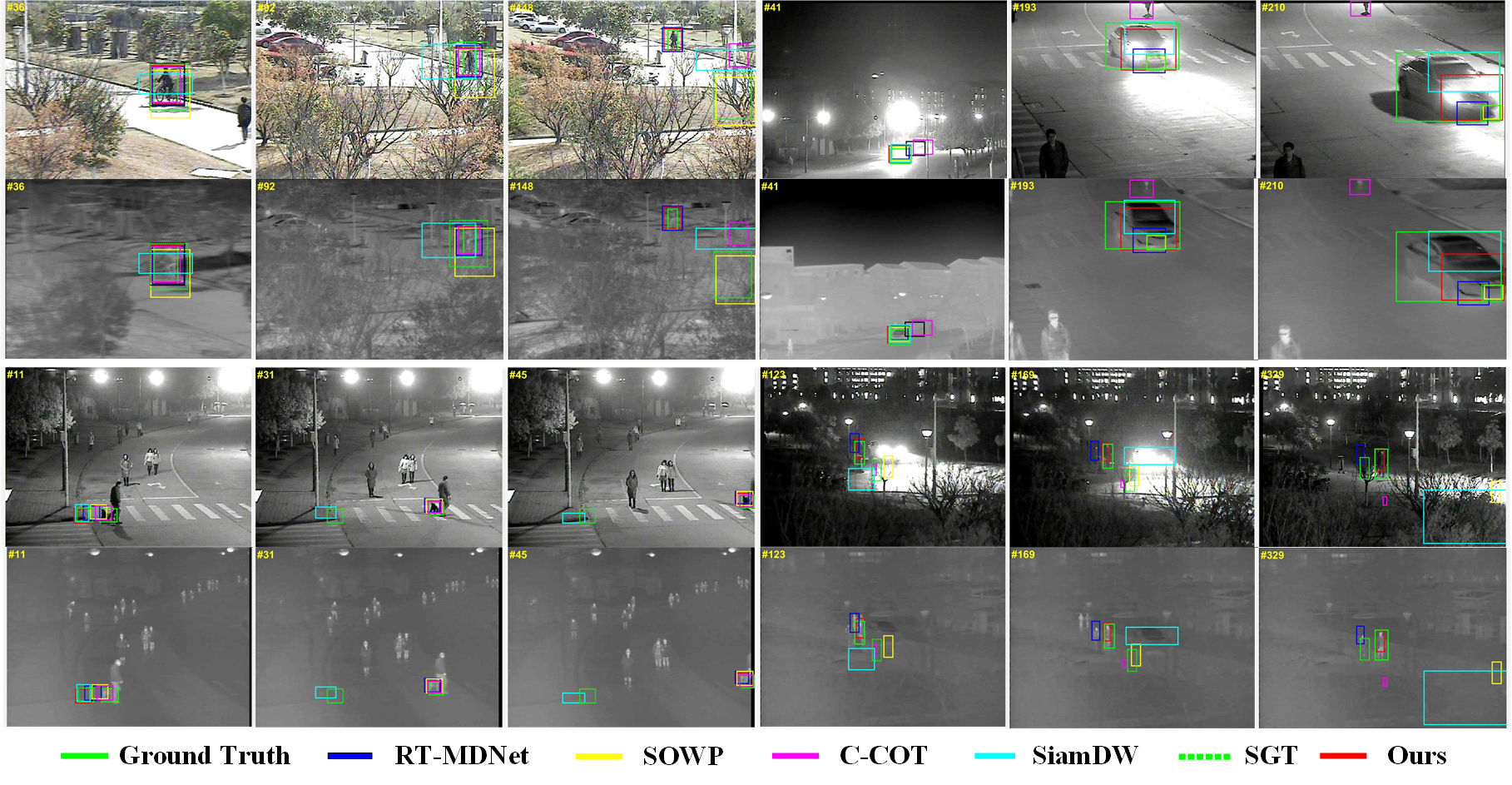}
\caption{Visualization of tracking results of compared state-of-the-art RGB-T trackers and our tracker. }
\label{resultsVIS}
\end{figure*}

\textbf{Tracking Results:} 
We visualize the tracking results of RT-MDNet, Ours, and Ground Truth with blue, red, and green rectangles respectively, as shown in Fig. \ref{resultsVIS}. We can find that our tracker is more robust than RT-MDNet to challenging factors like \emph{heavy occlusion}, \emph{abrupt change}, etc, due to the use of proposed modules. Compared with recent strong trackers like SiamDW and C-COT, our tracker can also attain better results as shown in Fig. \ref{resultsVIS}. These results all demonstrate the robustness of our proposed modules for RGB-T tracking.

\subsection{Discussion} 

The main contribution of this work is that we design a novel dynamic modality-aware filter generation network for accurate RGB-T tracking. We also validate that our modules can also work well with the attention mechanisms. The main difference between this work with Yang et al. \cite{yang2019learning} can be listed as: 
\emph{i}). Yang et al. use fixed filters which are learned offline for convolution on each test image, while we design a novel dynamic filter generation network that can predict various filters for different test images. 
\emph{ii}). Yang et al. use VGG as a backbone network and input a single video frame for global attention prediction, while we use resnet as the encoder and take continuous video frames for global attention generation. Our module considers the temporal information of video sequences with spatial and temporal RNN model. 
\emph{iii}). Yang et al. integrate their local and global attention modules into MDNet, while we develop our tracker based on real-time MDNet (i.e., RT-MDNet). We use spatial and channel attention for discriminative feature learning and can run faster than their tracker.

\begin{figure*} 
\center
\includegraphics[width=7in]{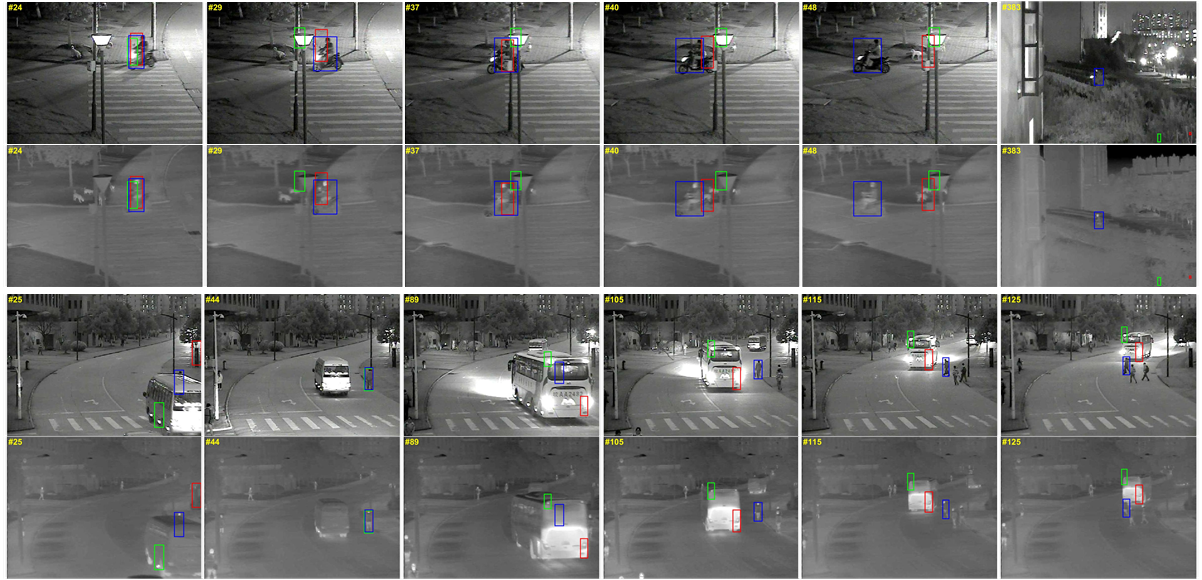}
\caption{Failed cases of our tracker. We denote the RT-MDNet, Ours and Ground Truth with green, red and blue rectangle respectively.}
\label{failedCase}
\end{figure*}

The aforementioned paragraphs demonstrate the effectiveness and advantages of our RGB-T tracker, however, our tracker also may fail to locate the target object in some challenging scenarios, as shown in Fig. \ref{failedCase}. Our global attention can help re-capture the target object to some extent, but when to switch to global search is still a challenging problem. Because we find that the response score can't always reflect the success/failure of tracking. In our future work, we will consider using reinforcement learning to learn a policy to adaptively switch between local and global search.

\section{Conclusion and Future Works}
In this paper, we propose a novel dynamic modality-aware filter generation module to boost the message communication between visible and thermal data. Given the image pairs as input, we first encode their features with the backbone network. Then, we concatenate these feature maps and generate dynamic context-aware filters with two independent networks. The visible and thermal filters will be used to conduct a dynamic convolutional operation on their corresponding input feature maps respectively. Inspired by residual connection, both the generated visible and thermal feature maps will be summarized with input feature maps. The augmented feature maps will be fed into the RoI align module to generate instance-level features for subsequent classification. Besides, we introduce the direction-aware RGB-T target driven attention for global proposal generation which can be used together with local search in the regular tracking-by-detection framework. Extensive experiments on three large-scale RGB-T tracking benchmark datasets validated the effectiveness of our proposed algorithm.

In our future works, we will focus on self-supervised learning to train our model with unlabeled RGB-T image pairs. Because only a small amount of labeled RGB-T tracking data can be used for the training which limits the applications of deeper neural networks.

\section*{Acknowledgements}
This work is jointly supported by Peng Cheng Laboratory Research Project (PCL2021A07), National Natural Science Foundation of China (61825101, 62027804, 62076004, 62102205, U20B2052), Natural Science Foundation of Anhui Province (2108085Y23), the Postdoctoral Innovative Talent Support Program (BX20200174), and the China Postdoctoral Science Foundation Funded Project (2020M682828).

\bibliographystyle{IEEEtran}
\bibliography{reference}

\end{document}